\documentclass[preprint,12pt, a4paper]{elsarticle}
\usepackage[T1]{fontenc}
\usepackage[utf8]{inputenc}
\usepackage{amssymb}
\usepackage{tikz}
\usepackage{lineno}
\usepackage{comment}
\usepackage{color}
\usepackage[normalem]{ulem}
\usepackage{listings}     
\usepackage{lstautogobble}  % Fix relative indenting
\usepackage{color}          % Code coloring
\usepackage{zi4}            % Nice font
\usepackage{subcaption}
\usepackage{url}
\usepackage{listings}
\definecolor{bluekeywords}{rgb}{0.13, 0.13, 1}
\definecolor{greencomments}{rgb}{0, 0.5, 0}
\definecolor{redstrings}{rgb}{0.9, 0, 0}
\definecolor{graynumbers}{rgb}{0.5, 0.5, 0.5}

\journal{SoftwareX}

\begin{document}

\begin{frontmatter}

\title{\emph{stream-learn} --- open-source \emph{Python} library for difficult data stream batch analysis}

\author{P. Ksieniewicz$^{[0000-0001-9578-8395]}$, P. Zyblewski$^{[0000-0002-4224-6709]}$}

\address{Wrocław University of Science and Technology\\Department of Systems and Computer Networks\\ wyb. Wyspiańskiego 27, 50-370 Wrocław}

\begin{abstract}
    \textsc{stream-learn} is a \emph{Python} package compatible with \emph{scikit-learn} and developed for the drifting and imbalanced data stream analysis. Its main component is a stream generator, which allows to produce a synthetic data stream that may incorporate each of the three main concept drift types (i.e. \emph{sudden}, \emph{gradual} and \emph{incremental drift}) in their \emph{recurring} or \emph{non-recurring} versions. The package allows conducting experiments following established evaluation methodologies (i.e. \emph{Test-Then-Train} and \emph{Prequential}). In addition, estimators adapted for data stream classification have been implemented, including both simple classifiers and \emph{state-of-art} chunk-based and online classifier ensembles. To improve computational efficiency, package utilises its own implementations of prediction metrics for imbalanced binary classification tasks.
\end{abstract}


\end{frontmatter}
%\linenumbers

\section{Motivation and significance}
\emph{Pattern recognition} research increasingly goes beyond the usual pattern of building classification models on stationary data sets and focuses on data stream processing where class distributions, and hence also decision boundaries, may change over time \cite{gama2009overview}. Such phenomenon is called the \emph{concept drift} \cite{Krawczyk2017} and causes the need to either update existing models or to replace them with completely new ones, depending on the characteristics of the changes taking place in the class distribution.

A completely different, but equally important issue is the \emph{imbalanced data} classification \cite{Krawczyk2016}. In such problems, the prior class  probabilities are uneven, and classic recognition models tend to have prediction bias towards the majority class. In the case of data streams, this issue may be further aggravated by dynamically changing the \emph{imbalance ratio} over time, where, even without \emph{concept drift}, consecutive models will indicate changes in the decision boundary. Despite the fact that real data streams often have a high and dynamically changing imbalance ratio, the number of scientific works combining the research trends of data streams classification and the skewed class distribution data analysis is still relatively small.

Research conducted in the "\emph{Imbalanced data stream classification algorithm}" project, aiming to deal with aforementioned problems, created a demand for a software that (\emph{i}) allows generating data streams with various properties (types of \emph{concept drift}, static and dynamic \emph{imbalance ratio}) in (\emph{ii}) class distributions based on quantitative features and for (\emph{iii}) proper evaluation of the proposed classification algorithms solving this kind of problems. An additional requirement was the compatibility with the estimators available in the \emph{scikit-learn} package \cite{scikit-learn}, being both an extremely popular research standard-software containing a huge amount of community-verified pattern recognition methods, as well as the \textsc{api} scheme for such libraries as \emph{imbalanced-learn} \cite{imblearn}, used to solve imbalanced classification problems or \emph{DESlib} \cite{cruz_deslib:2018}, used for dynamic classifier and ensemble selection.

The authors of the \emph{scikit-learn} package indicate in documentation the possibility of processing stream data\footnote{Section 8.1.1.3. Incremental learning: \url{https://scikit-learn.org/stable/modules/computing.html}} (which is also highlighted in the interfaces of some estimators that allow updating models using the \emph{partial-fit} method), but warn about the low efficiency of this type of approach when using online learning methods. This is due to the scripting nature of the \emph{Python} language, which is much more efficient at handling matrix operations (due to the \emph{numpy} package) than the nested loops being typical for low and middle-level languages.

The current solution for data stream processing using \emph{scikit-learn} interfaces is the \emph{scikit-multiflow} module \cite{skmultiflow}. It implements many of the methods present in the \textsc{moa} \cite{moa} package (a standard environment for data streams analysis in \emph{Java} that uses \textsc{weka} \cite{weka} interfaces), along with the evaluation methods. However, employing general processing idea from \emph{\textsc{moa}} software implies also the computationally inefficient paradigm of \emph{online processing} over \emph{batch processing}, where models do not receive individual samples, but their aggregated sets called \emph{data chunks}.

The \emph{stream-learn} module in the \emph{Python} language in accordance with the \emph{scikit-learn} \textsc{api} is the intended solution (i.e. batch-oriented data stream processing) to the problems mentioned above. As a base, it implements a data stream generator, based on the \emph{Madelon} \cite{Guyon2003} model used to generate static data in \emph{scikit-learn} and allows the development of both stationary and dynamic data streams, containing both \emph{concept} and \emph{prior class probabilities drifts}. It is supplemented with exemplary, simple stream classifiers (\emph{Accumulated Samples Classifier} and \emph{Sample Weighted Meta Estimator}), which may be used as the boilerplate for the users' solutions, and \emph{state-of-art} classifier ensembles (\emph{\textsc{sea}} \cite{Street2001}, \emph{OnlineBagging} \cite{Oza2005}, \emph{\textsc{oob}} \cite{Wang2015}, \emph{\textsc{uob}} \cite{Wang2015} and \emph{\textsc{wae}} \cite{Wozniak2013}). The package also implements evaluation metrics that are more computationally effective than those available in \emph{scikit-learn} and \emph{imbalanced-learn}. The element wrapping-up the package and allowing for conducting experiments is a pair of evaluators: \emph{Test-Then-Train} \cite{Gama2010} and \emph{Prequential} \cite{Gama2013}, in their batch variants.

The \emph{stream-learn} package is currently being developed at the \emph{Department of Systems and Computer Networks}, \emph{Wrocław University of Science and Technology}, surprisingly being also the place of its authors employment. It is used in scientific research related to the \emph{imbalanced data streams classification}. The articles created so far deal with topics such as the use of preprocessing in the incremental methods of imbalanced data stream classification \cite{Gulowaty2019}, the use of active learning techniques to reduce the number of patterns in streams \cite{KSIENIEWICZ201974} and exploring the possibilities of improving the classification of imbalanced data streams using the dynamic ensemble selection (\textsc{des}) \cite{Zyblewski2019,ecmliot}.

\section{Software description}
\subsection{Software Architecture}
\begin{figure}[!ht]
    \centering
    \includegraphics[width=13cm]{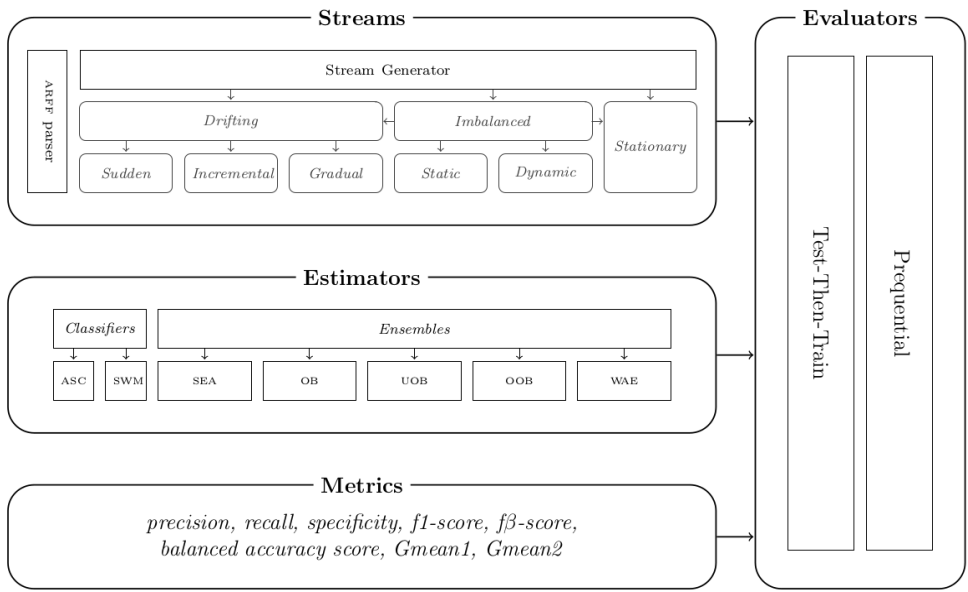}
    \caption{Overall schema of the software architecture.}
    \label{fig:architecture}
\end{figure}

The \emph{stream-learn} package is organised in five modules, responsible for (\emph{i}) data streams, (\emph{ii}) evaluation methods, (\emph{iii}) classification algorithms, (\emph{iv}) classifier ensembles and (\emph{v}) evaluation metrics. A general diagram of the project architecture is shown in Figure \ref{fig:architecture}.

The \emph{streams} module contains the \textsc{arff} file parser class, which is the standard format for serialising both real data streams and those generated, for example, by the \textsc{moa} software, as well as the \emph{StreamGenerator} class responsible for generating synthetic data streams. A more detailed description of the module can be found in Section 3.

The \emph{evaluators} module contains classes responsible for two main prediction measures estimation techniques on data streams, namely \emph{Test-Then-Train} and \emph{Prequential}, in their batch-based versions. The former one is based on separate windows known as data chunks, while the latter uses a sliding window as a forgetting mechanism. Both techniques, in each step, reevaluate existing classifiers.

Estimators can be found in the \emph{classifiers} and \emph{ensembles} modules, which contain the classifiers and \emph{state-of-art} classifier ensembles adapted for data stream classification, which can be used with implemented evaluators. A complete list of estimators includes:

\begin{itemize}
    \item Classifiers:
    \begin{itemize}
        \item \emph{Accumulated Samples Classifier} -- in each step concatenates all observed data chunks and fits the model on all encountered samples.
        \item \emph{Sample Weighted Meta Estimator} -- extends the partial-fit method of a given model by adding parameter allowing for weighting samples during classifier update.
    \end{itemize}
    \item Classifier ensembles:
    \begin{itemize}
        \item \emph{\textsc{sea}} -- \emph{Streaming Ensemble Algorithm}  \cite{Street2001}, trains a new base model on each incoming data chunk and adds it to the classifier pool but removes the worst model if the pool size is exceeded.
        \item \emph{OnlineBagging} \cite{Oza2005} -- uses the \emph{Poisson($\lambda= 1$)} distribution to update each base classifer with the appearance of a new instance. 
        \item \emph{\textsc{oob}} and \emph{\textsc{uob}} \cite{Wang2015} -- integrate resampling based on the $\lambda$ value with \emph{Online Bagging}.
        \item \textsc{wae} \cite{Wozniak2013} -- modifies the \emph{Accuracy Weighted Ensemble} (\textsc{awe}) \cite{Wang2003} by changing the weights calculation and classifier selection methods. 
    \end{itemize}
\end{itemize}

The \emph{metrics} module implements a wide range of evaluation measures for imbalanced binary classification \cite{Brzezinski2018}. The decision to prepare a new implementation was made due to the low computational efficiency of the metrics contained in existing packages. Module includes \emph{recall} \cite{Powers2011}, \emph{precision} \cite{Powers2011}, \emph{$f_\beta$ score} \cite{BaezaYates1999}, \emph{$f_1$ score} \cite{Sasaki2007}, \emph{balanced accuracy score} \cite{Brodersen2010,Kelleher2015} and two different definitions of \emph{geometric mean score} \cite{Barandela2003,Kubat1997}.

\subsection{Processing example}

The package structure will be much more transparent to the user after becoming familiar with the \emph{minimum processing example} included in the following subsection.

\subsubsection{Preparing experiments}

In order to conduct experiments, a declaration of four elements is necessary. The first is the estimator, which must be compatible with the \emph{scikit-learn} \textsc{api} and, in addition, implement the \emph{partial\_fit()} method, allowing  to re-fit the already built model. In the example, the standard \emph{Gaussian Naive Bayes} algorithm will be used:

\begin{lstlisting}[language=Python, caption=Declaring classifier for stream processing.]
from sklearn.naive_bayes import GaussianNB
clf = GaussianNB()
\end{lstlisting}

The next element is the data stream that will be analysed in processing. For the example purposes, it is a synthetic stream consisting of 100 chunks and precisely one concept drift. It will be prepared using the \emph{StreamGenerator} class of the \emph{streams} submodule:

\begin{lstlisting}[language=Python, caption=Generating a stream for processing.,firstnumber=3]
from strlearn.streams import StreamGenerator
stream = StreamGenerator(n_chunks=100, n_drifts=1)
\end{lstlisting}

The third requirement of the experiment is to specify the metrics used in the evaluation of the methods. In the example it is the \emph{accuracy score} metric available in \emph{scikit-learn} package and the \emph{precision} from the \emph{metrics} submodule:

\begin{lstlisting}[language=Python, caption=Declaring metrics for evaluation.,firstnumber=5]
from sklearn.metrics import accuracy_score
from strlearn.metrics import precision
metrics = [accuracy_score, precision]
\end{lstlisting}

The last necessary element of processing is the evaluator, i.e. the method of conducting the experiment. In the example the \emph{Test-Then-Train} paradigm, implemented in \emph{evaluators} submodule is chosen. It is important to note, that it is needed to provide the metrics used later in processing at the point of initialising the evaluator. In the case of none metrics given, it will use default pair of \emph{accuracy} and \emph{balanced accuracy} scores:

\begin{lstlisting}[language=Python, caption=Declaring evaluation method and pointing the metrics.,firstnumber=8]
from strlearn.evaluators import TestThenTrain
evaluator = TestThenTrain(metrics)
\end{lstlisting}

\subsubsection{Processing and understanding results}
Once all processing requirements have been met, the evaluation may be conducted. Calling the evaluator’s \emph{process} method, fed with given \emph{stream} and \emph{classifier} starts the evaluation:

\begin{lstlisting}[language=Python, caption=Running the experiment.,firstnumber=10]
evaluator.process(stream, clf)
\end{lstlisting}

The obtained results are stored in the \emph{scores} attribute of the evaluator. Printing the example's scores shows a three-dimensional \emph{numpy} array with dimensions \emph{(1, 99, 2)}:

\begin{itemize}
    \item The first dimension is the \emph{index of a classifier} submitted for processing. In the above example, only one model was used, but it is also possible to pass a tuple or a list of classifiers that will be processed in parallel.
    \item The second dimension specifies the \emph{instance of evaluation}, which in the case of \emph{Test-Then-Train} methodology directly means the index of the processed chunk (skipping the first chunk, which cannot be tested due to the lack of a model in the beginning of processing).
    \item The third dimension points the consecutive \emph{metric} used in the processing.
\end{itemize}

\begin{figure}[!ht]
    \centering
    \includegraphics[width=\textwidth, trim= 50 0 20 0]{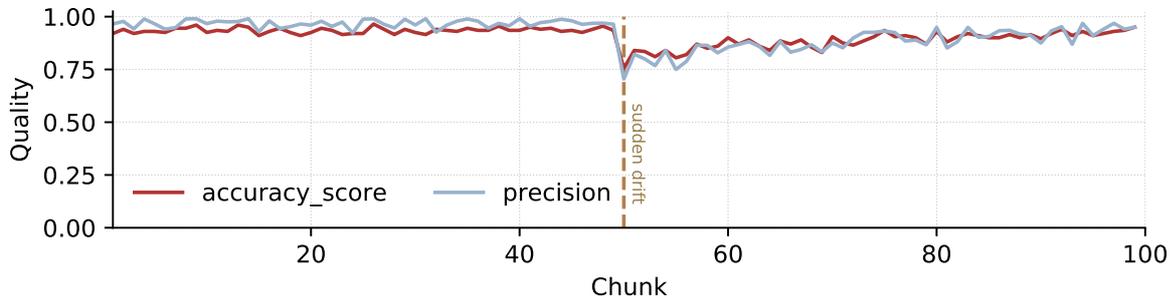}
    \caption{\emph{Accuracy score} and \emph{precision} achieved by a \emph{Gaussian Naive Bayes} classifier over the run of the synthetic data stream generated by the \emph{StreamGenerator} class.}
    \label{fig:simplest}
\end{figure}

Using this knowledge, it is finally possible to illustrate the results of a simple, exemplary experiment in the form of a plot illustrated in Figure \ref{fig:simplest}.

\section{Data stream generation}

A key element of the \emph{stream-learn} package is a generator that allows to prepare a replicable (according to the given \emph{seed}) classification dataset with class distribution changing over the course of a stream, with base concepts build on a default class distributions for the \emph{scikit-learn} package from the \emph{make\_classification()} function. These types of distributions try to reproduce the rules for generating the \emph{Madelon} set \cite{Guyon2003}. The \emph{StreamGenerator} is capable of preparing any variation of the data stream known in the general taxonomy of data streams.

\subsection{Stationary stream}

\begin{figure}[!ht]
    \centering
    \includegraphics[width=\textwidth]{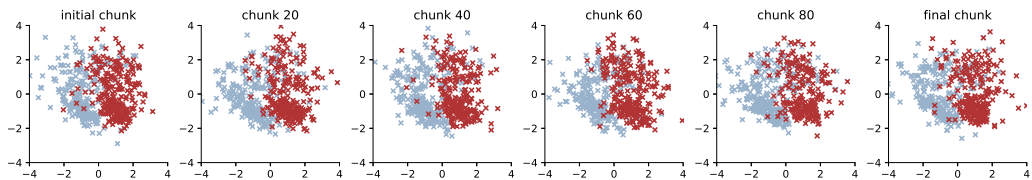}
    \caption{Scatter plots of selected chunks from a \emph{stationary data stream}.}
    \label{fig:stationary}
\end{figure}

The simplest variation of data streams are \emph{stationary streams}. They contain one basic concept, static for the whole course of the processing. Chunks differ from each other in terms of the patterns inside, but the decision boundaries of the models built on them should not be statistically different. This type of a stream may be generated with a clean generator call, without any additional parameters. Such a stream is illustrated in the Figure \ref{fig:stationary}, which contains the series of \emph{scatter plots} for a two-dimensional stationary stream with the binary problem.

What is important, contrary to a typical call to \emph{make\_classification()}, the \emph{n\_samples} parameter, determining the number of patterns in the set, is not specified here, but instead two new attributes of a data stream are provided:

\begin{itemize}
    \item \emph{n\_chunks} --- to determine the number of chunks in a data stream.
    \item \emph{chunk\_size} --- to determine the number of patterns in each data chunk.
\end{itemize}

Additionally, data streams may contain noise which, while not considered as a \emph{concept drift}, provides additional challenge during the data stream analysis and classifiers should be robust to it. The \emph{StreamGenerator} class implements noise by inverting the class labels of a given percentage of incoming instances in the data stream. This percentage can be defined by a \emph{y\_flip} parameter, like in standard \emph{scikit-learn} dataset generation call. If a single float is given as the parameter value, the percentage of noise refers to combined instances from all the classes, while if a tuple of floats is specified, the noise occurs within each class separately using the given percentages.

\subsection{Streams containing concept drifts}
The most commonly studied nature of data streams is their variability in time. Responsible for this is the phenomenon of the \emph{concept drift}, described in the introduction to this article. The \emph{stream-learn} package tries to meet the need to synthesise all the basic variations of this phenomenon (i.e. \emph{sudden}, \emph{gradual} and \emph{incremental drifts}).

\subsubsection{Sudden (Abrupt) drift}
This type of a drift occurs when the concept from which the data stream is generated is suddenly replaced by another one. Concept probabilities used by the \emph{StreamGenerator} class are created based on sigmoid function, which is generated using \emph{concept\_sigmoid\_spacing} parameter, which determines the function shape and how sudden the change of concept is. The higher the value, the more sudden the drift becomes. Here, this parameter takes the default value of 999, which allows for a generation of sigmoid function simulating an abrupt change in the data stream. Illustration of the \emph{sudden drift} is presented in the Figure \ref{fig:sudden}.

\subsubsection{Gradual drift}
Unlike \emph{sudden drifts}, gradual ones are associated with a slower change rate, which can be noticed during a longer observation of the data stream. This kind of drift refers to the transition phase where the probability of getting instances from the first concept decreases while the probability of sampling from the next concept increases. The \emph{StreamGenerator} class simulates \emph{gradual drift} by comparing the concept probabilities with the generated random noise and, depending on the result, selecting which concept is active at a given time. Illustration of the \emph{gradual drift} is presented in the Figure \ref{fig:gradual}.

\subsubsection{Incremental (Step-wise) drift}

\begin{figure}[!ht]
    \centering
    \begin{subfigure}[t]{\textwidth}
        \centering
        \includegraphics[width=\textwidth]{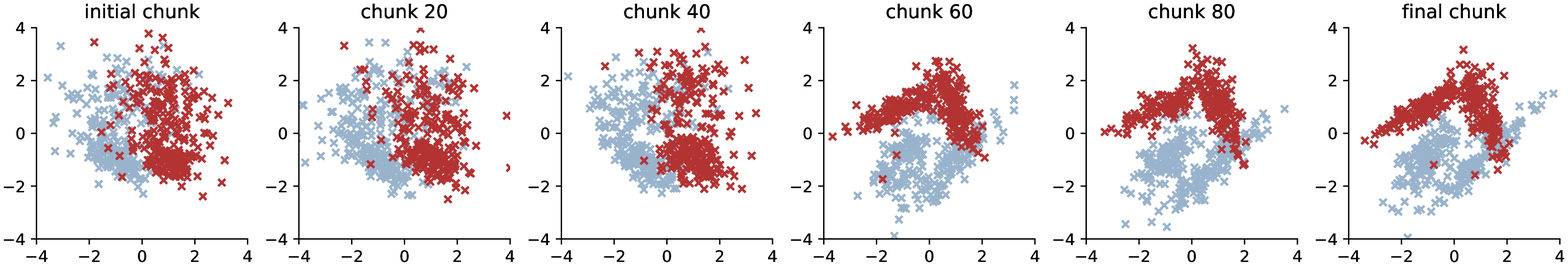}
        \caption{Data stream with sudden drift.}
        \label{fig:sudden}
    \end{subfigure}
    \begin{subfigure}[t]{\textwidth}
        \centering
        \includegraphics[width=\textwidth]{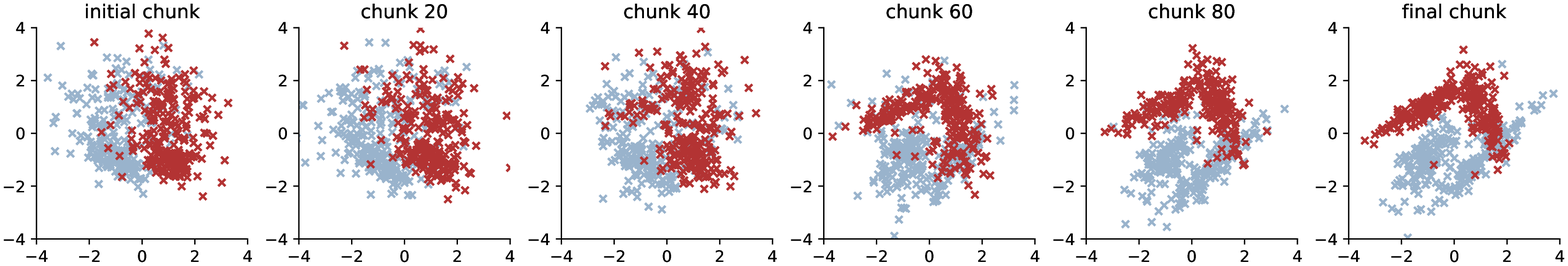}
        \caption{Data stream with gradual drift.}
        \label{fig:gradual}
    \end{subfigure}
    \begin{subfigure}[t]{\textwidth}
        \centering
        \includegraphics[width=\textwidth]{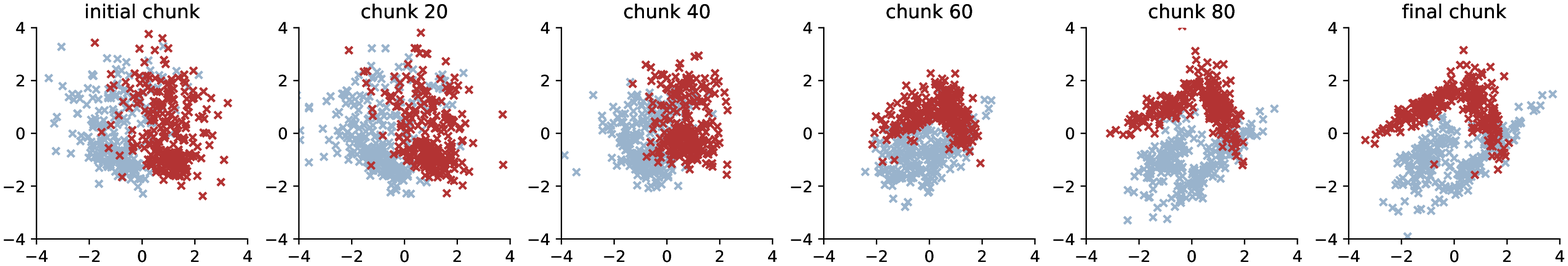}
        \caption{Data stream with incremental drift.}
        \label{fig:incremental}
    \end{subfigure}
    \caption{Changes in class distribution under each type of concept drift.}
\end{figure}

The \emph{incremental drift} happens when a series of barely noticeable changes in the concept used to generate the data stream occurs, in opposite of \emph{gradual drift}, which is mixing samples from different concepts without changing them. Due to this, the drift may be identified only after some time. The severity of changes, and hence the speed of transition of one concept into another is, like in previous example, described by the \emph{concept\_sigmoid\_spacing} parameter. Illustration of the \emph{incremental drift} is presented in the Figure \ref{fig:incremental}.

\subsubsection{Recurrent drift}
The cyclic repetition of class distributions is a completely different property of concept drifts. If after another drift, the concept earlier present in the stream returns, we are dealing with a \emph{recurrent drift}. We can get this kind of data stream by setting the \emph{recurring} flag in the generator. Illustration of the \emph{recurrent drift} is presented in the Figure \ref{fig:recurring}.

\subsubsection{Non-recurring drift}
\begin{figure}[!ht]
    \centering
    \begin{subfigure}[t]{\textwidth}
        \centering
        \includegraphics[width=\textwidth]{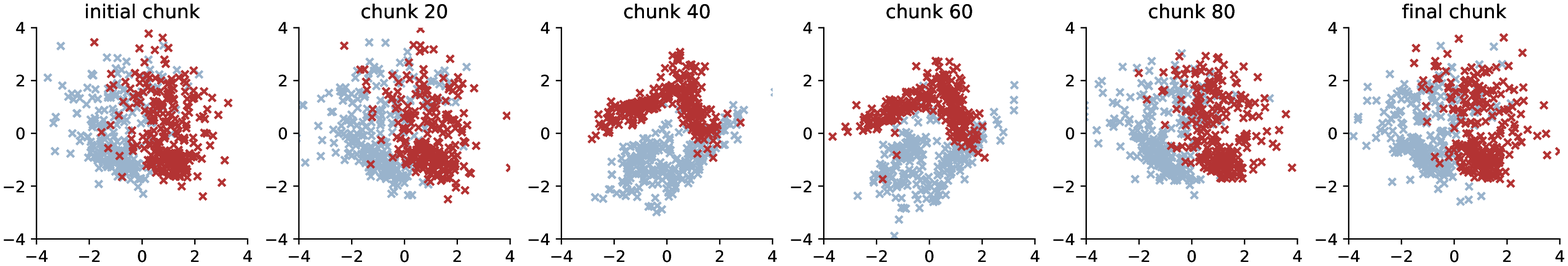}
        \caption{Data stream with recurring drift.}
        \label{fig:recurring}
    \end{subfigure}
    \begin{subfigure}[t]{\textwidth}
        \centering
        \includegraphics[width=\textwidth]{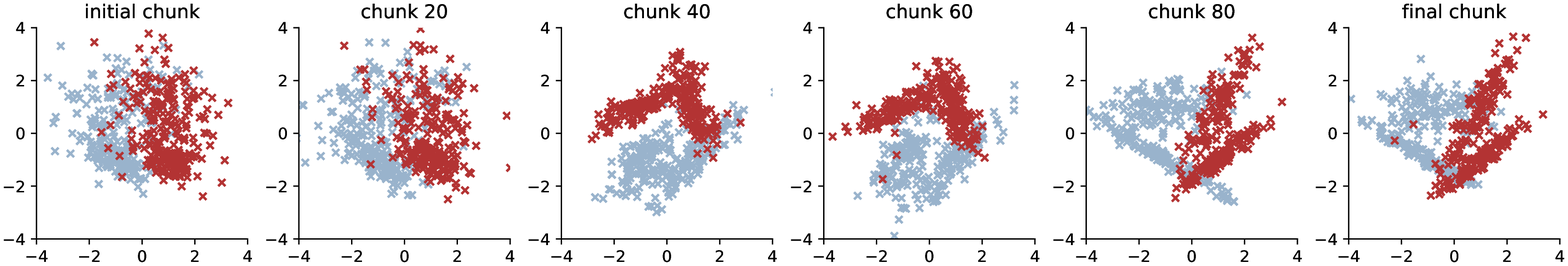}
        \caption{Data stream with non-recurring drift.}
        \label{fig:non-recurring-drift}
    \end{subfigure}
    \caption{Changes in class distribution under recurring and non-recurring concept drift.}
\end{figure}

The default mode of consecutive concept occurrences is a non-recurring drift, where in each concept drift a completely new, previously unseen class distribution is synthesised. Illustration of the \emph{non-recurring drift} is presented in the Figure \ref{fig:non-recurring-drift}.

\subsection{Class imbalance}
Another area of data stream properties, different from a \emph{concept drift} phenomenon, is the prior probability of problem classes. By default, a balanced stream is generated, i.e. one in which patterns of all classes are present in a similar number, like the stationary stream presented in Figure \ref{fig:simplest}.

\subsubsection{Stationary imbalanced stream}
The basic type of problem in which we are dealing with disturbed class distribution is a \emph{stationary imbalanced stream}, where the classes maintain a predetermined proportion in each chunk of data stream. To acquire this type of a stream, one should pass the \emph{list} to the \emph{weights} parameter of the generator (\emph{i}) consisting of as many elements as the classes in the problem and (\emph{ii}) adding up to one. Illustration of the \emph{stationary imbalanced} stream is presented in the Figure \ref{fig:static-imbalanced}.

\subsubsection{Dynamically imbalanced stream}
A less common type of \emph{imbalanced data}, impossible to obtain in static datasets, is \emph{data imbalanced dynamically}. In this case, the class distribution is not constant throughout the course of a stream, but changes over time, similar to changing the concept presence in gradual streams. To get this type of a data stream, a \emph{tuple} of three numeric values is passed to the \emph{weights} parameter of the generator:

\begin{itemize}
\item the number of cycles of distribution changes.
\item \emph{concept\_sigmoid\_spacing} parameter, deciding about the dynamics of changes on the same principle as in \emph{gradual} and \emph{incremental drifts}.
\item range within which oscillation is to take place.
\end{itemize}

\begin{figure}[!ht]
    \centering
    \begin{subfigure}[t]{\textwidth}
        \centering
        \includegraphics[width=\textwidth]{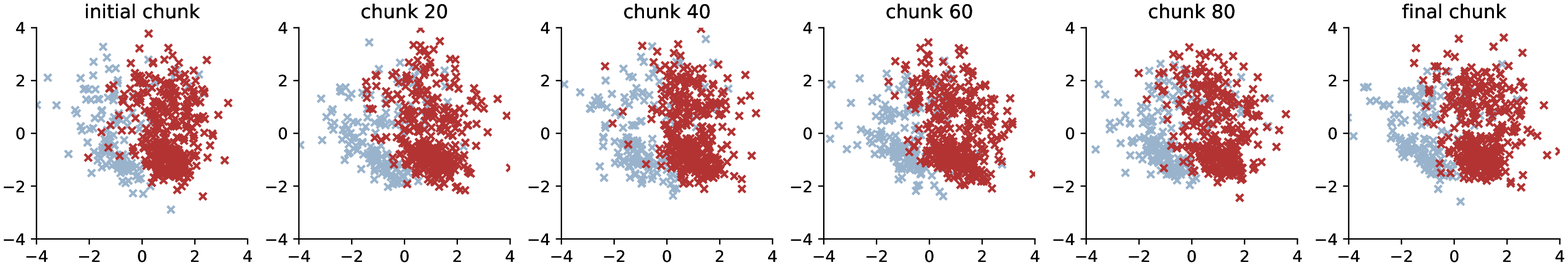}
        \caption{Statically imbalanced data stream.}
        \label{fig:static-imbalanced}
    \end{subfigure}
    \begin{subfigure}[t]{\textwidth}
        \centering
        \includegraphics[width=\textwidth]{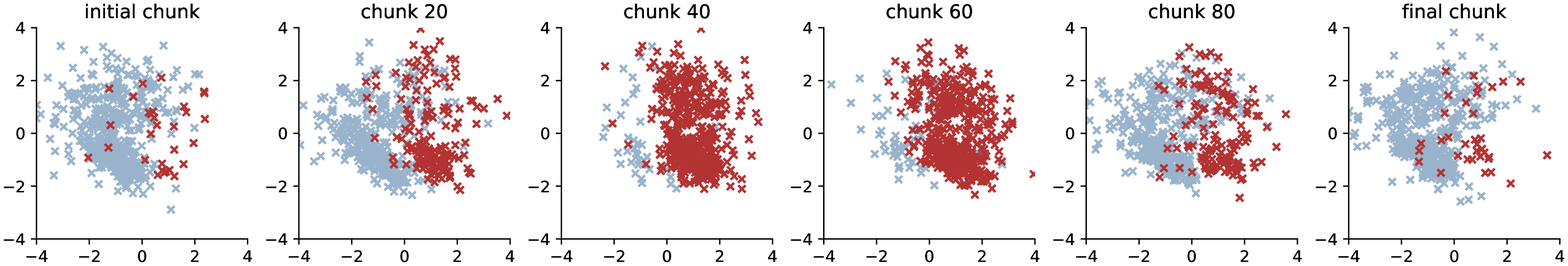}
        \caption{Dynamically imbalanced data stream.}
        \label{fig:dynamic-imbalanced}
    \end{subfigure}
    \begin{subfigure}[t]{\textwidth}
        \centering
        \includegraphics[width=\textwidth]{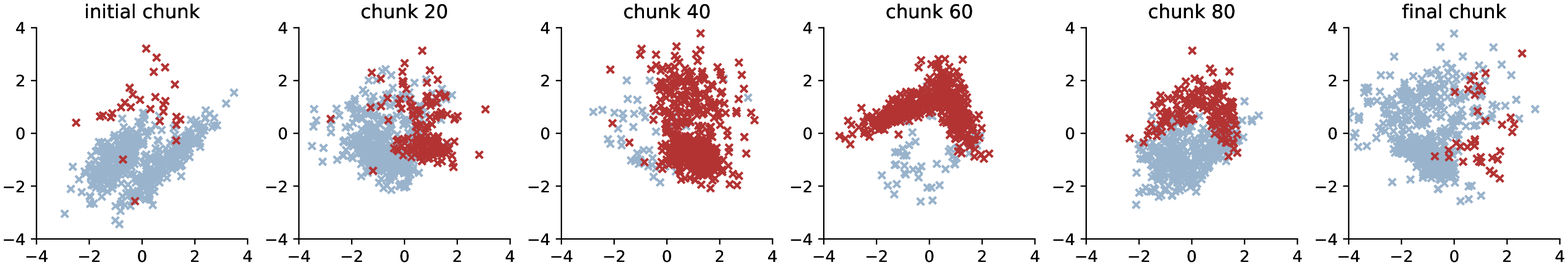}
        \caption{Dynamically Imbalanced Stream with Concept Oscillation (\textsc{disco}).}
        \label{fig:disco}
    \end{subfigure}
    \caption{Changes in class distribution under dynamically changing prior class probabilities (a,b) and concept drift paired with dynamic imbalance (c).}
\end{figure}

Illustration of the \emph{dynamically imbalanced stream} is presented in the Figure \ref{fig:dynamic-imbalanced}.

\subsection{Mixing drift properties}
Of course, when generating data streams, we do not have to limit ourselves to just one modification of their properties. One may easily prepare a stream with many drifts, any dynamics of changes, a selected type of drift and a diverse, dynamic imbalanced ratio. The last example of a data stream is such proposition, namely, \textsc{disco} (\emph{Dynamically Imbalanced Stream with Concept Oscillation}). Illustration of the \textsc{disco} stream is presented in the Figure \ref{fig:disco}.

\section{Impact}

The \emph{stream-learn} library allows to conduct research on the new algorithms for batch processing of data streams. Previous work employing it in a scientific process has already shown that it may be successfully used to verify the effectiveness of new methods against already established \emph{state-of-art} solutions implemented in accordance with the  \emph{scikit-learn} library. The most important area of research possible with its use is classification of data streams containing \emph{concept drifts} and/or \emph{dynamic class imbalance}. Thanks to the precise, replicable and user-friendly stream generation procedure, it also allows for a wide spectrum of \emph{drift detection} analyses, depending not only on types of drifts, but also on the dynamics of their changes. Finally, it also implements online bagging methods (\textsc{uob}, \textsc{oob}), which, to the knowledge of the authors, have not yet had open and stable implementation.

\section{Conclusions}
The \emph{stream-learn} package is a user-friendly and open source \emph{Python} library for difficult data stream classification. It allows generating streams with different characteristics, containing various types of concept drift and class-imbalance levels, including the possibility of drift in prior class probabilities. Additional modules allow conducting experiments on data streams using well-known estimation methodologies, implemented classifiers and classifier ensembles (some not present in any other \emph{Python} package). Its main idea is to let the user to immediately familiarize with the data stream classification task. 

The package has already been tested in the research process of preparing several scientific articles and it is an ideal tool for users who care about the simplicity of processing, ease of the use and integration with the \emph{scikit-learn} machine learning library.

\section*{Acknowledgements}

This work was supported by the Polish National Science Centre under the grant No. 2017/27/B/ST6/01325 as well as by the statutory funds of the Department of Systems and Computer Networks, Faculty of Electronics, Wroclaw University of Science and Technology.

\bibliographystyle{elsarticle-num} 
\bibliography{bibliography}

\section*{Required Metadata}
%\label{}

\section*{Current code version}
%\label{}

\begin{table}[!ht]
\caption{Code metadata (mandatory)}
\begin{tabular}{|l|p{6.5cm}|p{6.5cm}|}
\hline
\textbf{Nr.} & \textbf{Code metadata description} & \textbf{Please fill in this column} \\
\hline
C1 & Current code version & 0.8.5 \\
\hline
C2 & Permanent link to code/repository used for this code version & $https://github.com/w4k2/stream-learn$ \\
\hline
C3 & Legal Code License & GPL-3.0 \\
\hline
C4 & Code versioning system used & git \\
\hline
C5 & Software code languages, tools, and services used & python \\
\hline
C6 & Compilation requirements, operating environments \& dependencies & \\
\hline
C7 & If available Link to developer documentation/manual &  $https://w4k2.github.io/stream-learn/$ \\
\hline
C8 & Support email for questions & $pawel.zyblewski@pwr.edu.pl$\\
\hline
\end{tabular}
%\label{} 
\end{table}

\end{document}